%% file: main.tex
\documentclass[11pt]{article}

\usepackage{arxiv}

\usepackage[utf8]{inputenc}
\usepackage[T1]{fontenc}
\usepackage{microtype}

\usepackage{amsmath,amssymb,amsfonts}
\usepackage{newpxtext}   
\usepackage{newpxmath}   
\usepackage{bm}

\usepackage{booktabs}
\usepackage{multirow}
\usepackage{array}
\usepackage{float}   

\usepackage{graphicx}
\usepackage{tikz}
\usetikzlibrary{arrows.meta, positioning, shapes.geometric, calc, fit, backgrounds}
\usepackage{caption}
\DeclareCaptionLabelSeparator{vbar}{ $|$ }
\usepackage{subcaption}
\graphicspath{{figures/}}

\usepackage{enumitem}
\usepackage{xcolor}
\usepackage{url}

\usepackage[numbers,sort&compress]{natbib}
\usepackage{hyperref}
\usepackage[capitalise,noabbrev]{cleveref}

\input{macros}

\title{Geometry-Enhanced Portion Estimation for Multimodal LLMs}

\author{%
  Lin Liao \quad Peng Li \\[3pt]
  {\small \texttt{lin@8up.ai} \quad\quad \texttt{peng@8up.ai}}%
}

\date{July 2026}
\runningtitle{Geometry-Enhanced Portion Estimation for Multimodal LLMs}

\begin{document}
\maketitle

\input{abstract}

\input{intro}
\input{related}

\input{method}
\input{experiments}
\input{limitations}
\input{conclusion}

\bibliographystyle{plainnat}

\input{main.bbl}
\end{document}

%% file: macros.tex
\newcommand{\dino}{DINOv2\xspace}
\newcommand{\gemini}{Gemini-3.5-Flash\xspace}

\newcommand{\nfivek}{Nutrition5k\xspace}

\newcommand{\mae}{MAE\xspace}
\newcommand{\pmae}{PMAE\xspace}
\newcommand{\gram}[1]{#1\,\text{g}}

\newcommand{\ownm}{o}                  
\newcommand{\n}{n}                     
\newcommand{\p}{p}                     
\newcommand{\height}{h}               
\newcommand{\dens}{\rho}              
\newcommand{\massn}{m_n}             
\newcommand{\voln}{v_n}              

\providecommand{\percent}{\%}
\providecommand{\SI}[2]{#1\,#2}
\providecommand{\SIrange}[3]{#1--#2\,#3}

\usepackage{xspace}

\newcommand{\best}[1]{\textbf{#1}}

%% file: abstract.tex
\begin{abstract}
Image-based dietary assessment promises to replace costly, bias-prone manual
recalls, but portion estimation remains a major blocker.
Multimodal LLMs (MLLMs) recognize a wide range
of foods zero-shot in uncontrolled photos, yet they are weak at portion
estimation---a gap we measure across the current frontier (Gemini, GPT, and
Claude flagships alike). We present a method
that \emph{enhances a frozen, commercial MLLM with an accurate portion head}: a
small geometry-enhanced network on a frozen \dino backbone with a
\emph{structured softmax-ownership volume}, consuming the MLLM's per-food name,
bounding box, and density range---no depth sensor, no MLLM fine-tuning.
Evaluated fully open-vocabulary on three real-world benchmarks, the head cuts
per-food portion error by \textbf{\SIrange{33}{41}{\percent}} relative to the
MLLM alone, outperforms every flagship MLLM's direct estimates, and surpasses
each benchmark's originally published image-only model at its own reported
metric.
\end{abstract}

%% file: intro.tex
\section{Introduction}
\label{sec:intro}

Dietary assessment underpins nutrition research and public-health monitoring, but
the conventional instruments face a trade-off between burden, detail, and
accuracy~\citep{thompson2017dietary}. A 24-hour dietary recall captures fine-grained
intake but asks respondents to remember and report everything they ate, which is
burdensome and prone to recall bias. Food-frequency questionnaires are easier
to administer---respondents only report how often they eat items from a fixed
list---but sacrifice the detail needed to quantify specific foods and portions.
Image-based food recognition and portion estimation offer a scalable, objective
alternative that aims for detail without the respondent burden~\citep{boushey2017mobile}.
Existing methods, however, are developed on small, controlled datasets and do not
transfer to real life. Many tend to report only a single dish-level total
(mass, calories, or macronutrients); the lack of \emph{per-food identity and portion
size} makes them insufficient for dietary assessment and nutrition research, because
critical diet-quality scores such as the Healthy Eating Index (HEI) and
specific micronutrients (e.g., iron, sodium, vitamins) cannot be recovered from one aggregate.

Multimodal LLMs (MLLMs), trained on web-scale image--text data, recognize a wide
range of foods zero-shot in uncontrolled, real-world photos. This open-vocabulary
recognition is exactly what a real-life dietary tool requires, and MLLMs have
already been adopted in modern nutrition apps for image-based food recognition and
nutrition estimation. However, MLLMs have been shown to be weak at fine-grained
portion estimation, producing large discrepancies in energy and nutrient
estimates~\citep{li2024evaluating}.

We propose to \textbf{enhance a commercial MLLM with a dedicated portion
head}---keeping its open-vocabulary recognition while making the portion estimates
accurate enough for nutrition research on real-life images. Our head treats the MLLM
as a frozen upstream detector: for each detected food it returns a name, a bounding
box, and a plausible density range. The head then reasons jointly over all detected
foods and the image so the foods coordinate to partition shared regions of the plate.
From this it estimates each food's mass through structured prediction.
The design uses no depth sensor and does not fine-tune the MLLM itself.

\paragraph{Contributions.}
\begin{enumerate}[leftmargin=1.4em, itemsep=2pt, topsep=2pt]
  \item \textbf{A geometry-enhanced head that endows a frozen MLLM with accurate
    portion estimation.} Two components are central: an \emph{image$\times$food
    cross-attention transformer} that jointly reasons over all detected foods and
    the image so foods coordinate rather than double-count, and a \emph{structured
    softmax-ownership volume} that turns this into per-food mass. It substantially
    lowers per-food error relative to
    the MLLM alone, using no depth sensor and no MLLM fine-tuning.
  \item \textbf{A real-life system suitable for large-scale dietary assessment.} It
    extracts a broad range of foods from real-world meal photos and estimates
    per-food portion, which can be mapped to canonical food databases for
    high-fidelity nutrition information (e.g., macronutrients, micronutrients, and diet-quality scores).
\end{enumerate}

In the remainder of the paper we discuss related work, detail the model and its
training, and report its performance on three public food benchmarks.

%% file: related.tex
\section{Related Work}
\label{sec:related}

Image-based food recognition and portion estimation has received considerable attention
from both the nutrition-research and computer-vision communities.

\paragraph{Nutrition Research and Apps}
Image-assisted and image-based dietary assessment methods have been developed
over the last decade~\citep{boushey2017mobile}. In recent years, AI-based methods,
including convolutional neural networks (CNNs) and multimodal LLMs (MLLMs),
have been proposed and evaluated~\citep{shonkoff2023aibased, yan2025dietai24}.
Although image-based AI has shown great promise, large and inconsistent errors
remain a major barrier to its wide adoption in population-scale dietary assessment.
Consumer nutrition apps, such as MyFitnessPal and Noom, have adopted image-based
food tracking, but a substantial quality gap still separates them from
research-grade dietary assessment~\citep{li2024evaluating}.

\paragraph{Computer Vision}
Recovering mass or volume from a food image is the central hard sub-problem,
because a 2D photo discards the 3D information that determines how much food is
present. Early monocular approaches regress portion or energy directly from a
single RGB image, coupling food classification with portion regression in a
multi-task network~\citep{he2020multitask}. To restore
the missing geometry, a dominant line adds depth: \nfivek~\citep{thames2021nutrition5k}
supplies a sensor depth map per dish, and DPF-Nutrition~\citep{han2023dpf} predicts
depth from the RGB image and fuses it for nutrition estimation. A more recent line
makes geometry explicit, reconstructing a 3D food model or point cloud and inferring
scale from physical references or contextual objects~\citep{vinod2024portion3d,chen2026implicitscale}.
These geometry-based methods raise accuracy but depend on depth sensors, multi-view
capture, or fragile monocular 3D reconstruction, and operate over a closed food
vocabulary. In contrast, we keep a single RGB image with no explicit depth, learning
a structured volume from frozen appearance features while an MLLM supplies
open-vocabulary food identity.

%% file: method.tex
\section{Method}
\label{sec:method}

\begin{figure}[t]
  \centering
  \includegraphics[width=\linewidth]{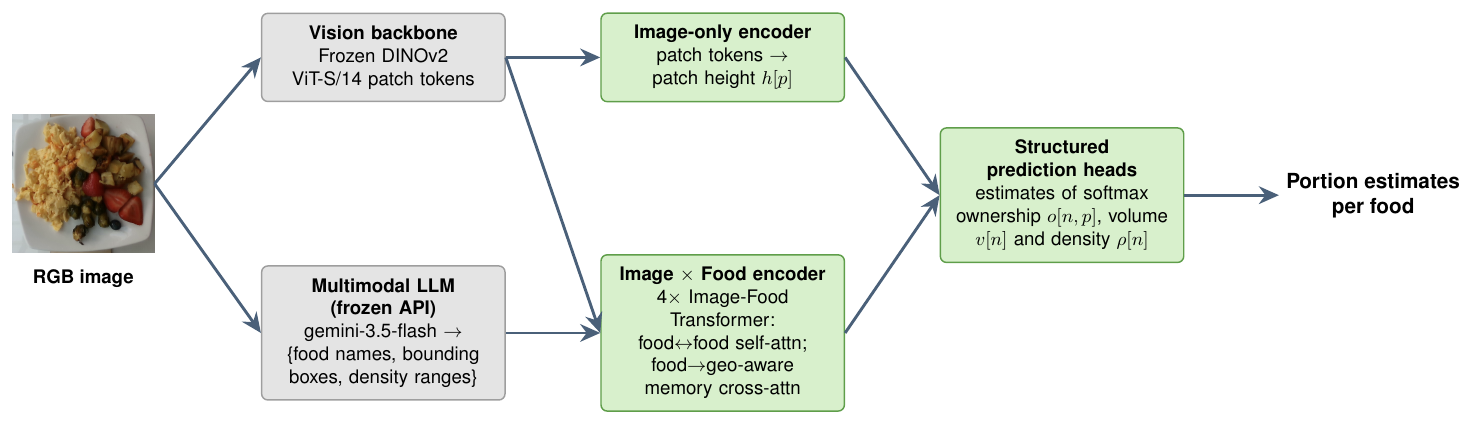}
  \caption{%
    \textbf{Geometry-enhanced MLLM architecture.} A frozen MLLM
    (\gemini) extracts per-food name, bounding box, and density range from the
    RGB image; a frozen \dino ViT-S/14 supplies patch tokens. The image$\times$food
    encoder and an image-only encoder feed structured heads that produce an
    exclusive per-patch ownership partition, a per-food density anchored to the
    MLLM range, and a per-patch height field. Volume estimate for food $\n$ is the sum
    across all patches $\p$:
    $\voln=\sum_p \ownm_{n,p}\,\height_p$ and per-food mass estimate is $\massn=\voln\cdot\dens_n$.}
  \label{fig:arch}
\end{figure}

Our system pairs a frozen commercial MLLM with a small, trainable
geometry-enhanced head. The MLLM supplies open-vocabulary semantics---per-food
name, bounding box, and density range---while the head supplies the geometry that
the MLLM lacks, turning appearance features into a calibrated per-food portion estimate.
\Cref{fig:arch} gives the overall data flow.

\subsection{Multimodal LLM and Vision Backbone}
\label{sec:vllm}
The MLLM is queried as a frozen API with no fine-tuning. We benchmarked five
flagship multimodal models and adopt \gemini~\citep{team2023gemini}, which pairs strong recognition
with the best direct portion accuracy among the flagships we benchmarked
(\cref{sec:mllm-choice}). For each of the $N$ detected foods it returns the
\textbf{name}, a \textbf{bounding box} enclosing all of that food's pieces, and a
\textbf{density range} $[\dens_{\min}, \dens_{\max}]$. The name is embedded with a
CLIP text encoder~\citep{radford2021clip} and is the dominant semantic signal; the box and density range
act as geometric and physical priors.

The vision backbone is a frozen \dino ViT-S/14~\citep{oquab2023dinov2}. We resize
every image---across all datasets---to a fixed \SI{336}{px} square; with the ViT's
\SI{14}{px} patch this yields a $24\times24=576$ patch-token grid
$\{x_p\}_{p=1}^{P}$ ($P{=}576$), cached for speed. Both encoders below read these same
frozen tokens; neither the backbone nor the MLLM is updated.

\subsection{Image\texorpdfstring{$\times$}{×}Food Encoder}
\label{sec:ixf}
\begin{figure}[t]
  \centering
  \includegraphics[width=0.72\linewidth]{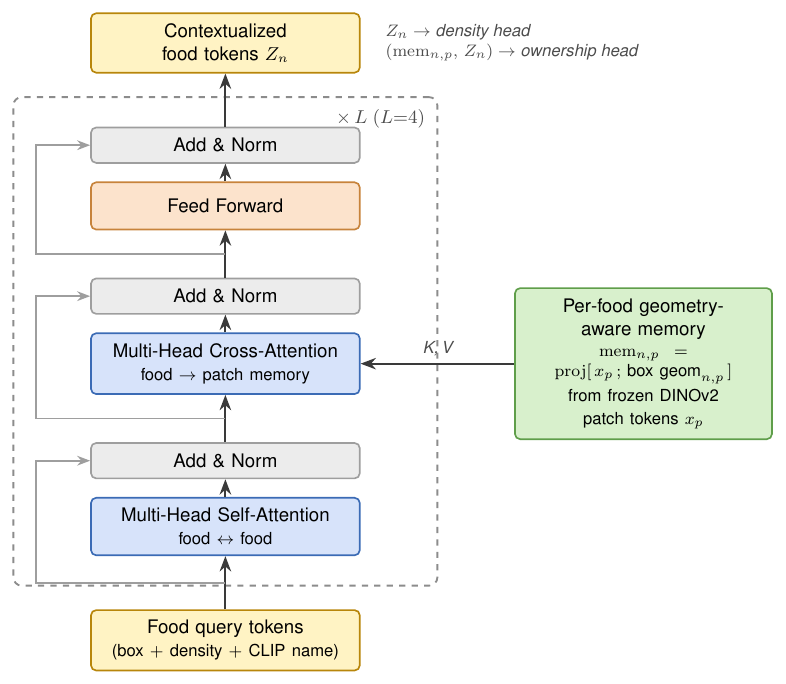}
  \caption{%
    \textbf{Image\texorpdfstring{$\times$}{×}Food encoder block.} Each of the $L{=}4$ layers is a
    Transformer decoder-style block: multi-head \emph{self}-attention over the
    food query tokens (food$\leftrightarrow$food), multi-head \emph{cross}-attention
    from the foods into the per-food geometry-aware patch memory
    $\mathrm{mem}_{n,p}$ (keys/values, built once from the frozen \dino patch
    tokens and per-food box geometry), and a feed-forward sublayer, each wrapped
    by a residual connection and layer normalization. The stack outputs
    contextualized per-food tokens $Z_n$: the density head reads $Z_n$ alone,
    while the ownership head scores each $(\mathrm{mem}_{n,p}, Z_n)$ pair.}
  \label{fig:ixf}
\end{figure}

This encoder produces, for every detected food, a contextualized food embedding
and, jointly with a per-food patch memory, the evidence the ownership and density
heads consume. Each of its layers is a Transformer decoder-style
block~\citep{vaswani2017attention}---self-attention over the food tokens,
cross-attention into the patch memory, and a feed-forward sublayer, without
causal masking since the foods form an unordered set (\cref{fig:ixf}).

\paragraph{Inputs.} (i)~$N$ \emph{food query tokens} $\{q_n\}$, each formed by
projecting the concatenation of that food's box features, density features, and
CLIP name embedding into the model width $d$---the shared hidden dimension of
all head components ($d{=}256$ throughout); (ii)~a \emph{per-food
geometry-aware memory}: for food $n$ and patch $p$, the entry
$\mathrm{mem}_{n,p}$ projects the frozen \dino patch token $x_p$, concatenated
with geometry features describing where $p$ falls relative to food $n$'s
bounding box, into width $d$ (plus a positional embedding). The memory is built
once from the $P$ frozen patch tokens and serves as the keys and values of
every layer's cross-attention.

\paragraph{Layer structure.} The queries pass through $L{=}4$ stacked layers.
Each layer applies three sublayers, each wrapped with a residual connection and
layer normalization:

\begin{itemize}[leftmargin=1.4em, itemsep=2pt, topsep=2pt]
  \item \textbf{Food$\leftrightarrow$food self-attention}, where the queries, keys,
    and values are all the food tokens. This lets foods coordinate so they can
    partition shared regions of the plate rather than each claiming the same
    pixels.
  \item \textbf{Food$\rightarrow$patch cross-attention}, where food $n$'s token
    attends into its own memory slice $\mathrm{mem}_{n,\cdot}$. Crucially the
    geometry is placed in the attention \emph{values}, not only as an additive
    attention bias: a bias would merely steer \emph{where} a food looks, whereas
    putting geometry in the values lets the model read it as \emph{content} and
    reason over it. We find this choice is load-bearing.
  \item \textbf{A position-wise feed-forward network} over each food token.
\end{itemize}

\paragraph{Outputs.} A contextualized per-food token $Z_n$. The density head
reads $Z_n$ directly; the ownership head scores the pair
$(\mathrm{mem}_{n,p},\,Z_n)$---how well food $n$'s contextualized token matches
its geometry-aware view of patch $p$---to produce the ownership logits of
\cref{sec:heads}. The memory is computed once by this encoder and reused
unchanged by the ownership head.

\subsection{Image-only Encoder}
\label{sec:height}
This encoder predicts a single positive scalar per patch---the ``pile-height''
field $\height_p$---from appearance alone, independent of which foods are present.

\paragraph{Structure.} The $P$ \dino patch tokens are linearly projected to width
$d$ and passed through a shallow \emph{patch self-attention} encoder (one
transformer layer, queries/keys/values all the projected patch tokens), so each
patch can use its spatial neighborhood---a patch in the middle of a tall pile
versus at a thin edge. A per-patch MLP then maps each token to one scalar, and a
softplus\footnote{$\mathrm{softplus}(x)=\log(1+e^{x})$, a smooth positive-valued
activation, so $\height_p>0$.} yields the height $\height_p>0$.

\paragraph{Input / output.} Input: the $P$ frozen patch tokens only (no food
tokens, no box, no name). Output: a field $\{\height_p\}$ on the $24\times24$ patch
grid. Because it does not depend on the food set, it is a reusable estimate of how
much the surface rises above the plate at each patch, shared by all foods when the
volume is integrated.

\paragraph{$\height_p$ is a latent height, not predicted depth.}
Although $\height_p$ plays the role that a depth map plays in depth-based portion
methods, it is deliberately \emph{not} predicted depth. It carries no depth
supervision: there is no sensor target and no depth loss anywhere in training, and
$\height_p$ is learned end-to-end \emph{only} through the downstream mass objective
(\cref{sec:training})---it becomes whatever makes the integrated volume, times
density, match the weighed grams. It is therefore a unitless, latent per-patch
weight rather than a calibrated metric surface, and it is defined at the coarse
$24\times24$ \emph{patch} resolution, not the dense per-pixel resolution of
monocular depth estimators. This is a design choice, not a limitation: explicit
single-view depth---sensor depth, pseudo-depth, and depth-supervised height
heads---is a tested negative in our setting, whereas a latent height learned under
mass supervision keeps the model depth-free while still recovering the missing
vertical dimension.

\subsection{Structured Prediction Heads}
\label{sec:heads}
The heads take three inputs from the encoders---the contextualized per-food tokens
$Z_n$ and the per-food, per-patch memory $\mathrm{mem}_{n,p}$ from the
image$\times$food encoder (\cref{sec:ixf}), and the per-patch height $\height_p$
from the image-only encoder (\cref{sec:height})---and turn them into a per-food
volume, density, and mass.

\paragraph{Ownership.} Ownership answers, for each patch, ``what fraction of this
patch belongs to each food?''. We first form a \emph{claim logit} for every
food--patch pair and a background logit for every patch,
\begin{equation}
  z_{n,p} = \mathrm{MLP}_{\mathrm{own}}\!\big([\,\mathrm{mem}_{n,p}\,;\,Z_n\,]\big),
  \qquad
  z_{\mathrm{bg},p} = \mathrm{MLP}_{\mathrm{bg}}(x_p),
\end{equation}
where $z_{n,p}$ scores how strongly food $n$ claims patch $p$---from the match
between its token $Z_n$ and the geometry-aware memory $\mathrm{mem}_{n,p}$---and
$z_{\mathrm{bg},p}$ scores how much patch $p$ is plate or empty. A softmax over the
$N$ foods \emph{and} background makes them compete for each patch:
\begin{equation}
  \ownm_{n,p}
  = \frac{\exp(z_{n,p})}
         {\exp(z_{\mathrm{bg},p}) + \sum_{n'=1}^{N}\exp(z_{n',p})},
  \qquad
  \sum_{n=1}^{N} \ownm_{n,p} + \ownm_{\mathrm{bg},p} = 1 .
  \label{eq:ownership}
\end{equation}
Hence $\ownm_{n,p}\in[0,1]$ is the \emph{share} of patch $p$ assigned to food $n$,
and the shares over $\{1,\dots,N,\mathrm{bg}\}$ partition each patch exactly. This
exclusive, soft partition is what makes the downstream volume well-behaved: because a
patch's area cannot be counted twice, foods that overlap must divide the shared
region between them, a larger food wins more patches, and background absorbs the
plate rather than inflating any food.

\paragraph{Density.} From each food token $Z_n$ a per-food log-density is predicted
and hinge-anchored to the MLLM's range $[\dens_{\min,n},\dens_{\max,n}]$, so
$\dens_n$ stays physically plausible while the model calibrates within the range.

\paragraph{Volume and mass.} The volume of food $n$ integrates the shared height
field over the patches it owns, and mass follows from density:
\begin{equation}
  \voln = \sum_{p} \ownm_{n,p}\,\height_p,
  \qquad
  \massn = \voln \cdot \dens_n,
  \qquad
  M_{\mathrm{dish}} = \sum_{n} \massn .
\end{equation}
The dish total is simply the sum of per-food masses. We deliberately omit a
learned dish-total residual head: it helps dish-total accuracy but overfits the
per-food objective.

\subsection{Training}
\label{sec:training}
Queries are \textbf{MLLM detections, not ground-truth foods}, and the same
detections are used at training and test time; ground-truth (GT) per-food mass is the only
supervision target, and no oracle food list is supplied (the per-dataset
recognition setup is described in \cref{sec:datasets}).

\paragraph{Match-and-mask supervision.} Because detections need not align with GT
ingredients, we align detections to GT foods one-to-one---by fuzzy name matching
on \nfivek, whose ingredients have no ground-truth boxes, and by token- and
descriptor-based name matching gated with box IoU where GT boxes exist (FPB,
NV-Real). We call the resulting scheme \emph{match-and-mask}: it is the
matching-based supervision of set-prediction models such as
DETR~\citep{carion2020detr}, except that unmatched queries are \emph{masked
out} of the loss rather than pushed toward a ``no object'' target. Concretely,
the per-food mass loss is applied \emph{only on matched detections}, while the
density hinge anchors every detection (matched or not) to a plausible density;
unmatched detections remain as no-loss queries whose appearance-grounded mass
is still summed at inference. The backbone stays frozen and
we optimize with AdamW; the structured-volume model is warm-started from a simpler
direct scalar-volume model (identical to the no-structured-ownership ablation in
\cref{sec:experiments}), since training the structured volume from scratch is less
stable.

\paragraph{Loss.} The per-food mass objective sums two Smooth-L1 terms with
complementary roles, plus the density hinge:
\begin{equation}
  \begin{aligned}
  \mathcal{L}
  = \sum_{n \in \mathrm{matched}}
      &\Big[ \mathrm{SmoothL1}\big(\log \massn, \log g_n\big)
          + \alpha \, \mathrm{SmoothL1}\big(\massn, g_n\big) \Big] \\
  &+ \lambda_{\dens} \sum_{n} \mathrm{hinge}\big(\dens_n, [\dens_{\min,n}, \dens_{\max,n}]\big),
  \end{aligned}
  \label{eq:loss}
\end{equation}
where $g_n$ is the GT mass, $\massn$ the predicted mass (\cref{sec:heads}), $\alpha$
the linear-gram weight, and $\lambda_{\dens}$ the density-hinge weight. The two mass
terms are complementary. The \emph{log-space} term is scale-invariant---it penalizes
\emph{relative} error, so a few-gram garnish and a large pile of rice contribute
comparable gradients despite spanning two orders of magnitude in grams, which keeps
optimization stable across the wide mass range. But scale-invariance also means it
under-penalizes \emph{absolute} error on large portions: a $20\%$ miss is only
\gram{1} on a \gram{5} food yet \gram{60} on a \gram{300} one, and the log term
weights the two equally. The \emph{linear-gram} term adds a penalty measured
directly in grams, restoring pressure on the heavy-portion under-prediction that the
log term alone tolerates. We keep $\alpha$ small---enough linear signal to correct
large misses, but not so much that the noisy heavy-portion tail dominates and
hurts the many small and medium foods.

%% file: experiments.tex
\section{Experiments}
\label{sec:experiments}

We show that a lightweight geometry head on a frozen commercial MLLM substantially
improves per-food portion accuracy over the MLLM's own gram estimates. We focus on
\emph{per-food} accuracy---identity-resolved mass---because that is what dietary
assessment needs. We evaluate on three real-world datasets (\cref{sec:datasets}), compare
per-food accuracy against the MLLM alone (\cref{sec:main-results}), compare our
portion estimates against each dataset's originally reported result (\cref{sec:dish-total}),
ablate the model's components on \nfivek (\cref{sec:ablations}), and compare
flagship MLLMs as direct portion estimators (\cref{sec:mllm-choice}).

\subsection{Datasets}
\label{sec:datasets}
We use three public datasets that provide real meal images with ground-truth
\emph{per-food} portions, so the method is exercised beyond any single source.
\begin{enumerate}[leftmargin=1.4em, itemsep=2pt, topsep=2pt]
  \item \textbf{\nfivek}~\citep{thames2021nutrition5k} is our primary controlled
benchmark: $\sim$5{,}000 cafeteria dishes with continuous, scale-weighed
per-ingredient masses, of which $\sim$3{,}500 have overhead captures from a fixed
RGB-D rig (the subset we use). Its single overhead viewpoint makes it the
canonical setting for the footprint-to-mass ambiguity we target; we evaluate on a
held-out test set of 502 dishes.
  \item \textbf{FPB}~\citep{sanatbyek2024fpb} (Food Portion Benchmark) is a large
set of 14{,}083 photos over 138 food classes of Central Asian cuisine with
manually annotated boxes and laboratory-measured component weights. Each dish is
photographed from \emph{multiple viewing angles} (a top-down view plus four side
angles); we train and test on this mixed-angle image pool rather than a single fixed
view, which stresses generalization to varied capture geometry and unfamiliar
cuisines. We evaluate on its 2{,}197-image test split.
  \item \textbf{NutritionVerse-Real (NV-Real)}~\citep{tai2023nutritionverse} is a
real consumer-photography set of 889 hand-collected smartphone images spanning 251
dishes, every ingredient individually weighed. Images are taken freehand from random
angles, with \emph{no fixed camera-to-food distance} and no scale reference, so it
reflects the realistic, uncontrolled phone-capture setting a deployed dietary tool
must handle. We evaluate on its 265-image test split.
\end{enumerate}

Although \nfivek ships with sensor depth, we deliberately ignore it: such
sensors are rarely available in practice, so all our results are image-only (RGB), no
depth.

\paragraph{Protocol.} On each dataset we train the same head---identical
architecture, losses, and hyper-parameters---on that dataset's
training split and evaluate on its held-out test split. Recognition is fully open-vocabulary on every
dataset: the MLLM names foods freely and never sees a class menu or a per-dish
food list, in training or testing. For FPB, whose dishes are regionally unfamiliar, we add a brief
cuisine context to the prompt (that the images show Central Asian / Kazakh /
Russian / international cafeteria food from a catering service in Kazakhstan).
This is a general hint, not a class menu or per-dish food list, so recognition stays
open-vocabulary.

\subsection{Main Results: Portion Estimates Versus MLLMs}
\label{sec:main-results}

We compare per-food portion accuracy---our model versus the MLLM's own gram
estimates---using per-food Mean Absolute Error (\mae) and percentage MAE (\pmae). Error
is measured only over matched detection--ground-truth food pairs $\mathcal{M}$ (see the
recognition note below):
\[
  \text{\mae} = \frac{1}{|\mathcal{M}|}\sum_{(\hat\imath,\,j)\in\mathcal{M}}
         \bigl|\hat m_{\hat\imath}-m^{\star}_{j}\bigr|,
  \qquad
  \text{\pmae} = 100\,\frac{\text{\mae}}{\overline{m^{\star}}},
\]
where $\hat m_{\hat\imath}$ is a predicted per-food mass, $m^{\star}_{j}$ the matched
ground-truth mass, and $\overline{m^{\star}}$ the mean of the matched ground-truth
masses $\{m^{\star}_{j}\}_{j\in\mathcal{M}}$ (unmatched foods enter neither term). For
statistical stability we run our model with three seeds and report the average
(\cref{tab:main}).

\input{data/tab_main.tex}

The geometry head cuts per-food portion error by \SIrange{33}{41}{\percent} relative to
the MLLM alone, and the gain holds across controlled (\nfivek, $-41\%$) and
uncontrolled real-world data (FPB, $-35\%$; NV-Real, $-33\%$).

\paragraph{Recognition coverage (mass-weighted).}
Because per-food \mae is defined only on matched foods, missed and spurious
detections are excluded; recognition quality therefore bounds \emph{coverage} but
affects what the portion head sees but not the head's per-food accuracy on
matched pairs. Measured as mass-weighted recall,
our open-vocabulary recognition covers 87\% on \nfivek, 88\% on
NV-Real, and 64\% on FPB. FPB's lower coverage is a
recognition-\emph{identity} limit, not a weighing one: without a menu the MLLM
\emph{describes} unfamiliar regional dishes instead of naming them (chak-chak
becomes ``puffed rice cereal''; orama nan becomes ``steamed dumplings'').

\subsection{Portion Estimates Versus the Original Papers}
\label{sec:dish-total}

The three source papers report accuracy at different granularities, so we compare
our model to each on the metric that paper reports (\cref{tab:dish}), all in the
image-only, no-depth setting. The \nfivek and NV-Real papers predict only a
\emph{dish total}---an image backbone (InceptionV2 and Inception-ResNet,
respectively) regresses total dish mass directly, with no per-food breakdown---so
for these we compare our dish total, the sum of our per-food masses, against their
reported total. The FPB paper is different: its YOLOv12 model
regresses a weight per detected food item and reports a per-food \mae, so
for FPB we compare at the per-food level instead.
We stay depth-free and never regress the total directly; for reference the \nfivek
paper's depth/volume model reaches \gram{29.4}, but it uses depth.

\input{data/tab_dish.tex}

For context, \citet{thames2021nutrition5k} report human portion estimation
error on a small subset of \nfivek dishes: non-nutritionists average
\SI{53}{\percent} per-image percentage error, while trained nutritionists
average \SI{41}{\percent}.

\subsection{Ablations}
\label{sec:ablations}
We remove one component at a time from the full model and report per-food \mae,
broken down by single-food and multi-food dishes (\cref{tab:ablation}); this
ablation study is conducted on \nfivek only. \textbf{(1)~No MLLM
context:} zeros the per-food name, density, and box geometry, leaving only \dino patches
and a generic query, isolating the value of the MLLM priors. \textbf{(2)~No structured
ownership:} replaces the ownership$\cdot$height volume with direct mass regression, testing
whether the exclusive patch partition is load-bearing on mass. \textbf{(3)~No image-only
height:} sets $\height_p{=}1$, collapsing volume to footprint area.

\input{data/tab_ablation.tex}

The MLLM per-food context is by far the most load-bearing component: removing name, box,
and density more than \emph{doubles} per-food error (\gram{16.7}$\to$\gram{34.4},
$+106\%$). For both single-food and multi-food dishes, the errors go up greatly without
per-food context.
The error is especially large on multi-food dishes, because without per-food
priors the model can only regress toward the average mass of the foods on the plate.
The structured softmax-ownership partition
($+\gram{1.6}$, $+10\%$) and the learned image-only height field ($+\gram{1.2}$, $+7\%$)
each add a smaller, consistent gain across both single- and multi-food dishes.

\subsection{Which MLLM? A Flagship Comparison}
\label{sec:mllm-choice}
Our head is trained on \gemini outputs but its architecture is
MLLM-agnostic, so we benchmarked five flagships as \emph{direct}
portion estimators---each recognizes foods and estimates per-food grams itself,
open-vocabulary, with the same decomposition prompt and matcher---on 100
multi-food dishes ($\geq$2 significant foods, the hard regime of
\cref{tab:ablation}) from the \nfivek test set.
Each model runs at the minimum reasoning effort it allows, with temperature 0
where supported; all models were accessed through their public APIs in July
2026.

\input{data/tab_mllm.tex}

Results are shown in \cref{tab:mllm}. Three observations.
\textbf{(1) Recognition is broadly comparable.} On these hard cases,
mass-weighted coverage clusters at 61--72\% across all five models, with
Claude-Fable-5 highest at 71.6\%.
\textbf{(2) Portion estimation splits the field.}
The Gemini models and GPT-5.5 sit at \gram{24.1}--\gram{26.7}, while both
Claude models are near \gram{36}; the Claude models under-estimate large piles
less but pay with larger errors on ordinary portions---a different bias
profile, not a uniformly weaker one.
\textbf{(3) A small trained head beats every flagship.} Our portion estimation head,
fed \gemini recognition, reaches \gram{16.5} per-food \mae on the same 100 dishes---32\% below the best
flagship direct estimate.

%% file: data/tab_main.tex
\begin{table}[H]
  \centering
  \caption{\textbf{Per-food MAE/PMAE: our model vs.\ the MLLM alone.} Lower is better.
  Our model is fully open-vocabulary on every dataset.}
  \label{tab:main}
  \small
  \setlength{\tabcolsep}{5pt}
  \begin{tabular}{l c c c}
    \toprule
    Dataset & MLLM alone & Ours & MAE reduction \\
    \midrule
    \nfivek
      & \gram{28.1} / \SI{36.1}{\percent}
      & \best{\gram{16.7}} / \SI{21.4}{\percent}
      & \best{$-41\%$} \\
    FPB
      & \gram{66.9} / \SI{31.8}{\percent}
      & \best{\gram{43.6}} / \SI{19.3}{\percent}
      & \best{$-35\%$} \\
    NV-Real
      & \gram{41.0} / \SI{33.1}{\percent}
      & \best{\gram{27.4}} / \SI{21.0}{\percent}
      & \best{$-33\%$} \\
    \bottomrule
  \end{tabular}
\end{table}

%% file: data/tab_dish.tex
\begin{table}[H]
  \centering
  \caption{\textbf{Portion estimates vs.\ the original papers} (image-only, no
    depth). Each dataset is compared on the granularity its paper reports:
    \emph{dish total MAE} for \nfivek and NV-Real, and \emph{per-food MAE} for FPB.
    Lower is better.}
  \label{tab:dish}
  \begin{tabular}{l l c c}
    \toprule
    Dataset & Metric & Original paper & Ours \\
    \midrule
    \nfivek
      & dish total MAE
      & \gram{40.4} / \SI{18.8}{\percent}
      & \best{\gram{36.6}} / \SI{18.4}{\percent} \\
    NV-Real
      & dish total MAE
      & \gram{115.9} / \SI{27.2}{\percent}
      & \best{\gram{68.6}} / \SI{16.1}{\percent} \\
    FPB
      & per-food MAE
      & \gram{90.95} / \SI{40.3}{\percent}
      & \best{\gram{43.6}} / \SI{19.3}{\percent} \\
    \bottomrule
  \end{tabular}
\end{table}

%% file: data/tab_ablation.tex
\begin{table}[H]
  \centering
  \caption{\textbf{Ablations on \nfivek.} One component removed at a time from the
    full model; per-food \mae (grams, three-seed average) on the held-out test, broken down by single-food
    dishes, multi-food dishes, and all dishes (total), each with $\Delta$ vs.\ the full model. Lower is
    better.}
  \label{tab:ablation}
  \small
  \setlength{\tabcolsep}{3.5pt}
  \begin{tabular}{l cc cc cc}
    \toprule
    & \multicolumn{2}{c}{Single} & \multicolumn{2}{c}{Multi} & \multicolumn{2}{c}{Total} \\
    \cmidrule(lr){2-3}\cmidrule(lr){4-5}\cmidrule(lr){6-7}
    Variant & MAE & $\Delta$ & MAE & $\Delta$ & MAE & $\Delta$ \\
    \midrule
    Full model
      & \best{\gram{14.2}} & ---
      & \best{\gram{17.1}} & ---
      & \best{\gram{16.7}} & --- \\
    (1) No MLLM context (no name/density/box)
      & \gram{20.6} & $+\gram{6.4}$
      & \gram{36.7} & $+\gram{19.6}$
      & \gram{34.4} & $+\gram{17.7}$ \\
    (2) No structured ownership (direct mass regr.)
      & \gram{16.4} & $+\gram{2.3}$
      & \gram{18.6} & $+\gram{1.5}$
      & \gram{18.3} & $+\gram{1.6}$ \\
    (3) No image-only height ($\height_p{=}1$)
      & \gram{15.4} & $+\gram{1.2}$
      & \gram{18.3} & $+\gram{1.2}$
      & \gram{17.9} & $+\gram{1.2}$ \\
    \bottomrule
  \end{tabular}
\end{table}

%% file: data/tab_mllm.tex
\begin{table}[H]
  \centering
  \caption{\textbf{Flagship MLLMs as direct portion estimators.} Each model
    recognizes foods and estimates per-food grams directly (open-vocabulary,
    same decomposition prompt and matcher) on 100 multi-food dishes.
    Mass cov.\ = mass-weighted recall; higher is better. Per-food \mae: lower is
    better. Our model inherits \gemini recognition through the pipeline's
    extraction prompt (name, box, density---no gram estimation), so its
    coverage differs slightly from the Gemini-3.5-Flash row, whose detections
    come from the direct-estimation prompt: same model, different prompt.}
  \label{tab:mllm}
  \small
  \setlength{\tabcolsep}{4.5pt}
  \begin{tabular}{l c c}
    \toprule
    Model & Mass cov. & Per-food \mae \\
    \midrule
    Gemini-3.5-Flash
      & 68.9\% & \gram{24.1} / \SI{34.7}{\percent} \\
    Gemini-3.1-Pro
      & 68.6\% & \gram{25.8} / \SI{36.6}{\percent} \\
    GPT-5.5
      & 61.4\% & \gram{26.7} / \SI{38.7}{\percent} \\
    Claude-Fable-5
      & \best{71.6\%} & \gram{35.7} / \SI{51.3}{\percent} \\
    Claude-Opus-4.8
      & 66.8\% & \gram{36.2} / \SI{52.9}{\percent} \\
    \midrule
    Our model (head on \gemini)
      & 71.1\% & \best{\gram{16.5}} / \SI{23.7}{\percent} \\
    \bottomrule
  \end{tabular}
\end{table}

%% file: limitations.tex
\section{Limitations}
\label{sec:limitations}

Four limitations bound the current system, in decreasing order of impact.
First, \textbf{single-view portion is fundamentally ambiguous}: a footprint does
not determine pile height, so the heavy-portion tail is systematically
under-predicted. This is a property of the input, not of the head---we observe
it consistently across datasets.
Second, \textbf{image-only recognition cannot capture invisible foods}:
ingredients that leave little or no visual trace---cooking oil, added sugar,
salt, sauces mixed into a dish---are missed by any image-based recognizer, yet
they carry real mass and a disproportionate share of the nutrients dietary
assessment cares about.
Third, \textbf{the head is trained per dataset}: the architecture, losses, and
hyper-parameters are identical everywhere, but each benchmark uses its own
weights, and cross-dataset transfer of a single universal weigher has not been
demonstrated.
Fourth, \textbf{we validate mass, not nutrients}: the per-food
masses are designed for downstream nutrient conversion, but that conversion
is not evaluated here.

%% file: conclusion.tex
\section{Conclusion and Future Directions}
\label{sec:conclusion}

Food \emph{recognition} is now largely an MLLM capability; food \emph{weighing}
is not---every flagship we tested under-performs at portion estimation, and
larger reasoning models do not close the gap. Our results show this gap is
closed not by scale but by a small, task-specific geometry head: consuming only
the MLLM's structured output and frozen \dino features, it cuts per-food portion
error by \SIrange{33}{41}{\percent} relative to the MLLM alone, beats every
flagship's direct estimates, and surpasses each benchmark's originally published
image-only model---fully open-vocabulary, depth-free, and without touching the
MLLM. Recognition and weighing thus decouple cleanly: the MLLM supplies identity
and priors, the head supplies geometry.

The end goal is not a better plate estimator but a replacement for the 24-hour
dietary recall---the instrument nutrition research actually runs on---and
framing future work against that target makes the remaining gaps concrete.
\textbf{(1)~From served portion to consumed intake.} A recall measures what was
\emph{eaten}; an image measures what was \emph{served}. Closing the difference
requires paired before/after capture (or short video) with leftover
subtraction---a natural extension of a per-food weigher, since intake is the difference of
two portion estimates over matched foods. \textbf{(2)~From visible foods to
complete intake.} A recall interviewer \emph{probes}: cooking oil, sugar in
coffee, sauces mixed in, the composition of a casserole. No image model can see
these; the MLLM's dietary knowledge, used conversationally (brief follow-up
questions, recipe inference from dish identity), can substitute for the
probe---turning the invisible-ingredient limit from a silent bias into a bounded,
queryable one. \textbf{(3)~From plates to days.} A recall covers all eating
occasions; a deployed system must handle missed captures, snacks, and drinks
across a full day with one universal weigher (the fact that our identical
architecture and hyper-parameters already work across three diverse datasets
suggests one is trainable) and graceful degradation when images are absent.
\textbf{(4)~From benchmark agreement to clinical validity.} Replacing the
recall ultimately requires validation the way recalls themselves were
validated: against recovery biomarkers and controlled-feeding ground truth, on
nutrient intakes rather than grams, across diverse cuisines---where our
coverage results show current MLLMs still under-serve underrepresented foods.
Per-food mass estimation of the accuracy demonstrated here is, we believe, the
enabling component that makes this agenda realistic.